\documentclass[11pt,a4paper]{article}
\usepackage[hyperref]{ms}
\usepackage{times}
\usepackage{latexsym}
\usepackage{graphicx}

\usepackage{color}
\usepackage{url}

\aclfinalcopy 

\begin{document}
\title{Several Experiments on Investigating  Pretraining and Knowledge-Enhanced Models for Natural Language Inference}


\author{ Tianda Li 
\\ ECE, Queen's University
\\\tt tianda.li@queensu.ca
\\
\bf Quan Liu
\\iFLYTEK Research
\\ \tt quanliu@ustc.edu.cn
\\
\bf Zhigang Chen 
\\  iFLYTEK Research 
\\\tt zgchen@iflytek.com
\And
Xiaodan Zhu 
\\ ECE, Queen's University
\\ \tt zhu2048@gmail.com 
\\\bf Qian Chen 
\\University of Science and Technology of China
\\ \tt cq1231@mail.ustc.edu.cn
\\\bf Si Wei 
\\iFLYTEK Research 
\\\tt siwei@iflytek.com
 }

\date{December 2018}

\maketitle
\begin{abstract}
  Natural language inference (NLI) is among the most challenging tasks in natural language understanding.
  Recent work on unsupervised pretraining that leverages unsupervised signals such as language-model and sentence prediction objectives has shown to be very effective on a wide range of NLP problems. It would still be desirable to further understand how it helps NLI; e.g., if it learns artifacts in data annotation or instead learn true inference knowledge. In addition, external knowledge that does not exist in the limited amount of NLI training data may be added to NLI models in two typical ways, e.g., from human-created resources or an unsupervised pretraining paradigm. We runs several experiments here to investigate whether they help NLI in the same way, and if not, how?
  
\end{abstract}

\section{Introduction}
Modelling informal reasoning in natural language is a very challenging problem in natural language understanding. The recent availability of relatively large annotated datasets ~\cite{DBLP:conf/emnlp/BowmanAPM15,williams2017broad} have made it feasible to train complex
natural language inference (NLI) models that need to estimate a large number of parameters, including neural network models. Such models have shown to achieve the state-of-the-art performance~\citep{DBLP:conf/emnlp/BowmanAPM15,DBLP:conf/acl/BowmanGRGMP16,DBLP:conf/eacl/YuM17,DBLP:conf/emnlp/ParikhT0U16,DBLP:conf/coling/ShaCSL16,DBLP:conf/acl/ChenZLWJI17,DBLP:journals/corr/abs-1801-00102}. 
However, many recent research efforts \citep{DBLP:journals/corr/abs-1805-02266,DBLP:journals/corr/abs-1809-02719,DBLP:journals/corr/abs-1806-00692,DBLP:journals/corr/abs-1805-01042} have also observed that complex models may achieve better evaluation scores by over-exploiting the artifacts in data construction rather than modelling real NLI semantics.


As the recent advance in learning representation for natural language,  unsupervised pretraining that leverages large unannotated data using language-model or sentence prediction objectives have shown to be effective on a wide range of NLP tasks. It would, however, be desirable to understand if such models leverage artifacts in data construction or actually help learn inference related semantics. 

From a more general viewpoint, pretrained models, in parallel to those explicitly exploring human authorized semantics~\citep{chen2018neural}, incorporate external knowledge into NLI by learning from unlabelled data. 



We run several experiments showing that the recent pretraining schema do help learn NLI-related semantics to achieve better NLI prediction. The experiments also reveal that external knowledge obtained from pretraining and human authorized resources complement each other, suggesting potential benefits of incorporating the latter to the pretraining-finetuning schema.





\section{Related Work}

\paragraph{Unsupervised Pretraining}
Unsupervised pretraining has recently shown to be very effective in improving the performances of a wide range of NLP tasks. Feature-based and finetune-based models are two main strategies for using them in downstream tasks. Feature-based models such as ELMo~\citep{DBLP:journals/corr/abs-1802-05365} incorporate pretrained representation into task-specific models. As the approach keeps the original task-specific models intact, it can be seen as a powerful feature extractor for specific tasks.

Finetune-based approaches such as Generative Pretrained Transformer (GPT)~\citep{radford2018improving} and BERT~\citep{devlin2018bert}, however, 
pretrain a model on unannotated data and then finetune the same architecture and use it on different downstream tasks. 

Pretraining has been used in NLI~\citep{DBLP:journals/corr/abs-1802-05365,devlin2018bert} and shown to improve performance on many tasks. As complex models often overfit to class-conditional idiosyncrasies in the existing NLI datasets,
in this paper we run several experiments to understand if the pretraining schema helps learn true NLI related semantics to help achieve better NLI prediction.




 
 
\paragraph{Evaluation in Natural Language Inference} Previous research has paid attention to judge whether existing NLI systems have learned NLI-related semantics or explored the regularities existing in the data that are not relevant to NLI.
For example~\citep{DBLP:journals/corr/abs-1803-02324} found that in the construction phase of SNLI datasets, 
due to the strategies that the human subjects create hypotheses,
the distributions of words among NLI categories are different, which may not be relevant to NLI semantics. 
~\citet{DBLP:journals/corr/abs-1806-00692} pointed out that 
complex machine learning models have the capacity to exploit such artifacts. This characteristic allows the models to achieve better performance on many benchmark datasets.

To investigate this,
different methods have been proposed. For example,
~\citet{DBLP:journals/corr/abs-1809-02719} use a \textit{swapping} evaluation method,
by switching a premise and its hypothesis to change the distribution of words to test the robustness of a model.
Also,
efforts have also been made to propose new test dataset~\citep{DBLP:journals/corr/abs-1805-02266}.
In the test set,
premises are taken from the SNLI training set and 
for each premise, hypotheses of different inference categories (i.e., entailment, neural, and contradiction) are generated by replacing a single word in premise sentence. In addition, the stress test proposed by ~\citet{DBLP:journals/corr/abs-1806-00692} construct stress test dataset based on existing MultiNLI corpus~\citep{williams2017broad}.


\section{Experiment Set-Up}
\label{sec:setup}
\paragraph{Models} Our experiments use the the following models:Note that for all the previously published models, we either used the original code provided by its authors, or if not available, our implementation achieved a performance comparable to that reported in the original papers. 

\begin{itemize}
    \setlength\itemsep{-0.5em}
    \item BERT~\citep{devlin2018bert}
    \item DenseNet+DynAtt: DenseNet plus Dynamic Self Attention. This is a state-of-the-art sentence-embedding-based model proposed by~\citep{DBLP:journals/corr/abs-1808-07383}.
    \item DenseNet+MultiHeads: This is a model proposed in this paper to further investigate DenseNet+DynAtt, by replacing its top dynamic self-attention layer with multi-head attention used by~\citep{DBLP:journals/corr/abs-1806-09828}, to help observe the role of dynamic attention in NLI.
    \item LSTM+DynAtt: This is a model proposed in this paper to further investigate DenseNet+DynAtt, by replacing its lower DenseNet layer with LSTM as in ~\citep{DBLP:journals/corr/abs-1806-09828}, to observe the role of DenseNet in NLI.
    \item ESIM~\citep{DBLP:conf/acl/ChenZLWJI17}.~\footnote{ESIM is a strong NLI baseline. We used the source code made available at \textit{https://github.com/lukecq1231/nli}. The code can run efficiently and has been adapted for summarization~\citep{ChenZLWJ16ijcai} and question-answering tasks~\citep{Zhang:qa:2017}.}
    \item ESIM+ELMo~\citep{DBLP:journals/corr/abs-1802-05365}
    \item GenPool: Generalized Pooling~\citep{DBLP:journals/corr/abs-1806-09828}
    \item GenPool+ELMo: This is a model proposed in this paper to add ELMo to GenPool to observe the effect of ELMo on GenPool.        
    \item GPT~\citep{radford2018improving}
    \item KIM~\citep{chen2018neural}
\end{itemize}


\begin{table*}[!ht]
\centering
\small
\renewcommand{\arraystretch}{1.2}
\begin{tabular}{lll|lll|lll|lll}
\hline
\multicolumn{3}{l|}{} &\multicolumn{3}{l|}{Glockner Entailment (982)} & \multicolumn{3}{l|}{Glockner Neutral (47)} & \multicolumn{3}{l}{Glockner Contradiction (7164)} \\
\hline
\multicolumn{3}{l|}{Models} & P & R & F1 & P & R & F1 & P & R & F1 \\ \hline

\multicolumn{3}{l|}{BERT} & .800 & .980 & \textbf{.880} & .050 & .260 & .080 & .990 & .930 & \textbf{.960}  \\ 
\multicolumn{3}{l|}{DenseNet+DynAtt} & .394 & .908 & .550 & .002 & .021 & .003 & .979 & .732 & \textbf.834 \\ 
\multicolumn{3}{l|}{DenseNet+MultiHeads} & .274 & .914 & .421 & .000 & .000 & .000 & .977 & .655 & .784 \\ 
\multicolumn{3}{l|}{LSTM+DynAtt} & .214 & .990 & .352 & .008 & .043 & .014 & .992 & .472 & .640 \\ 
\multicolumn{3}{l|}{ESIM} & .226 & .994 & .368 & .006 & .064 & .011 & .992 & .466 & .634 \\ 
\multicolumn{3}{l|}{ESIM+ELMo} & .238 & .982 & .383 & .006 & .085 & .012 & .992 & .488 & .654 \\ 
\multicolumn{3}{l|}{GenPool} & .163 & .993 & .280 & .005 & .021 & .007 & .994 & .275 & .431 \\ 
\multicolumn{3}{l|}{GenPool+ELMo} & .162 & .996 & .279 & .010 & .043 & .016 & .996 & .272 & .428 \\ 

\multicolumn{3}{l|}{GPT} & .735 & .849 & \textbf{.788} & .013 & .511 & .027 & .994 & .726 & \textbf{.839} \\ 
\multicolumn{3}{l|}{KIM} & .450 & .973 & \textbf{.615} & .017 & .128 & .029 & .991 & .790 & \textbf{.879} \\ \hline
 
\end{tabular}

\caption{Model performances on the Glockner dataset. 
}
\label{glockner}
\end{table*}

\paragraph{Data} We use both existing data and that we further annotated in our experiments. The existing data include SNLI~\citep{DBLP:conf/emnlp/BowmanAPM15}, MultiNLI~\citep{williams2017broad}, the Glockner dataset~\citep{DBLP:journals/corr/abs-1805-02266}, and the stress test dataset~\citep{DBLP:journals/corr/abs-1806-00692}. The first two are used for training models and the last two for testing. We will discuss the subset that we further annotate later in the experiment result section.




\section{Experiment Results and Discussion}
\subsection{Pretraining Helps Learn Inference Knowledge}
As discussed above, the improvement in NLI modelling can be due to the models' sophisticated capability in capturing annotation artifacts introduced in data construction. While the recent pretrained models achieved impressive performance on a wide range of NLP tasks, including NLI, we show that they do learn NLI related semantics. 


\paragraph{The Glockner and Swapping Test}
~\citep{DBLP:journals/corr/abs-1805-02266} take the premises from SNLI and uses lexical replacement to generate hypotheses for different NLI categories (i.e., entailment, contradiction, and neutral) by replacing a single word in the premise and asking human subjects to confirm the replacement and the resulting inference relationships are correct.

We first perform tests on the Glockner dataset and then annotate a subset to provide some more insights. Table~\ref{glockner} shows the results on the original Glockner test data, with all models trained on SNLI. Note that the Glockner dataset has 982, 47, and 7164 test cases for entailment, neural, and contradiction, respectively, as marked in the table.



In general we can see that GPT, BERT, and KIM outperform the other models. As the Glockner test set is constructed to specifically focus on lexical semantics in NLI, the results suggest that GPT, BERT, and KIM capture lexical level inference knowledge. As a comparison, we can see that ESIM, which relies on the SNLI training data only, shows an inferior performance on this dataset. Note that KIM incorporates WordNet knowledge, while GPT and BERT learn external knowledge automatically from unannotated data. 

DenseNet+DynAtt is a state-of-the-art model and its performance is close to that of GPT in the contradiction category. We investigate the roles of the DenseNet and DynAtt in NLI by replacing these main components to create DenseNet+Multiheads and LSTM+DynAtt models, as explained in Section~\ref{sec:setup}. 
The results show that the DenseNet and DynAtt jointly work very well to contribute to the final performance, as the performance drops significantly when either of the components is replaced with those in other state-of-the-art models.


\citet{DBLP:journals/corr/abs-1809-02719} proposed an interesting idea to evaluate whether a model learns NLI-related semantics or explore statistics irrelevant to NLI by using a \textit{swapping} strategy.
The method switches a premise with its hypothesis to test a model: if a model learns entailment, it will have a large performance drop when tested on the swapped pairs.

We note that caution should be exercised as the swapping evaluation may not be conclusive if entailment sentence pairs (e.g., those in the SNLI and the Glockner data set) contain also paraphrases (premise entails hypothesis and vice versa).    


For this reason, we further manually annotated the 982 entailment pairs in the Glockner dataset, which contains only 32 non-paraphrase entailment sentence pairs. We perform the state-of-the-art models on these pairs and their swapping versions. Again, as discussed in~\citep{DBLP:journals/corr/abs-1809-02719}, a larger decrease of accuracy before and after the swapping corresponds to a better entailment model.

In Table~\ref{diff}, we see the differences of accuracies for the models. The performances of KIM, GPT, and BERT all have the largest differences before and after the swapping, indicating their better performance on this entailment test. Again, in KIM, different semantic relations between word pairs are incorporated from WordNet, while the experiment suggests that the pretrain-finetune schema can learn such knowledge (e.g., here hypernym replacement) for NLI. 
In the table we also include the differences between models on the paraphrasing subset. As discussed, the results are not conclusive. We list it here as one needs to be careful when applying the swapping test on the existing NLI data set.

\begin{table}[t!]
\small
\renewcommand{\arraystretch}{1.4}
\begin{tabular}{lll}
\hline
                    & Strict entailment & Paraphrasing \\ \hline
BERT                & \textbf{.969}     & -.007       \\
DenseNet+DynAtt     & .593             & -.019       \\
DenseNet+MultiHeads & .625             & -.023       \\
LSTM+DynAtt         & .437             & -.001         \\
ESIM                & .25             & -.154       \\
ESIM+ELMo           & .219             & -.001       \\
GenPool             & .000             & .001        \\
GenPool+ELMo        & .125             &  .027        \\
GPT                 & \textbf{.844}    & -.101       \\
KIM                 & \textbf{.781}    & -.006       \\ \hline
\end{tabular}
 \caption{Accuracy differences of each model before and after swapping, on the strict entailment subset and paraphrasing subset of the Glockner dataset, respectively.}
 \label{diff}
\end{table}

\subsection{Investigating External Knowledge from Different Sources}
Although the current NLI training datasets~\cite{DBLP:conf/emnlp/BowmanAPM15} are much larger than what were available previously, the amount of NLI knowledge that can be learned is still limited.
As suggested in ~\citep{DBLP:journals/corr/abs-1805-02266}, external knowledge from WordNet can significantly improve the performance of NLI prediction on the Glockner dataset. Also as discussed above, a NLI model may benefit from external knowledge in two typical ways, e.g., from human authorized sources ~\citep{chen2018neural} or from large unannotated data, e.g., via unsupervised pretraining~\citep{devlin2018bert,radford2018improving}.
In order to find out how NLI systems benefit from different external knowledge sources and whether they complement each other, we carry out further experiments. Specifically, we perform stress test~\citep{DBLP:journals/corr/abs-1806-00692} on KIM, GPT, BERT, and ESIM.

\begin{table}[h!]
\centering
\small
\renewcommand{\arraystretch}{1.1}
\begin{tabular}{lllll}
\hline
 & BERT   & GPT    & KIM    & ESIM   \\ \hline
BERT             & .561 & .580 & \textbf{.652} & .616 \\
GPT              &        & .304 & .543 & .457 \\
KIM              &        &        & .491 & .552 \\
ESIM             &        &        &        & .320 \\ \hline
\end{tabular}
\caption{Oracle accuracy of merging two models in stress test on the in-domain set.
  }
\label{matched}
\end{table}

\begin{table}[h!]
\centering
\small
\renewcommand{\arraystretch}{1.1}
\begin{tabular}{lllll}
\hline
 & BERT   & GPT    & KIM    & ESIM   \\ \hline
BERT                & .482 & .504  & \textbf{.580} & .539\\
GPT                 &        & .251 & .482 & .378 \\
KIM                 &        &        & .426 & .488 \\
ESIM                &        &        &        & .265 \\ \hline
\end{tabular}
\caption{Oracle accuracy of merging two systems in stress test on the out-of-domain test set.}
\label{mismatched}
\end{table}

The stress test proposed in ~\citep{DBLP:journals/corr/abs-1806-00692} was constructed in three categories:
competence, distraction, and noise test.
We use the competence test data to evaluate the models' ability to understand antonym relations in NLI, as the relationship between the other two categories and NLI prediction is less straightforward. 
We perform KIM, GPT, BERT, and ESIM on the competence test and derive Table~\ref{matched}, in which the cross of a row (e.g., BERT) and a column (e.g., KIM) is the oracle accuracy of merging the two models (i.e., 0.652). That is, for each test case, if any of the two models makes the correct predication, we regard the answer to be right. In this way, we show every single model's performance (along the diagonal) and the oracle performance of merging two models.


\begin{table*}[t!]
\renewcommand{\arraystretch}{1.5}
\small
\begin{tabular}{cl}
\hline
Example        &Sentences \\ \hline
1              &P: There are two people \textbf{inside}, and two men outside, a cafe; with a tv on in the background.    \\
              &H: There are two people \textbf{outside}, and two men outside, a cafe; with a tv on in the background. 

                                                      \\ 
                                                      \hline

\end{tabular}
 \caption{Examples on which KIM is right but BERT is wrong.}
 \label{KIM_sample}
\end{table*}

\begin{table*}[t!]
\renewcommand{\arraystretch}{1.5}
\small
\begin{tabular}{lcl}
\hline
Categories                    &Example                         & Sentences          \\ \hline
\uppercase\expandafter{\romannumeral1}               
           & 2  &P: Yellow banners with a black lion print are hung across some trees in a \textbf{sun-lit} neighborhood. \\
             &                                         &H: Yellow banners with a black lion print are hung across some trees in a \textbf{moon-lit} neighborhood.   

                                                      \\ 
& 3  &P: A young boy takes the first step onto \textbf{Mars}.    \\
&     &H: A young boy takes the first step onto \textbf{Earth}.   

                                                      \\ 
                                                      \hline
                  

\uppercase\expandafter{\romannumeral2} 

         & 4   &P: A Vietnamese woman gives a manicure	a \textbf{South} Korean woman gives a manicure.   \\
         &   &H: A Vietnamese woman gives a manicure	a \textbf{North} Korean woman gives a manicure.    
        \\
        & 5    &P: An \textbf{Indian} man is perching on top of a wall with a hammer and chisel.    \\
        &     &H: An \textbf{Indonesian} man is perching on top of a wall with a hammer and chisel.

                                                      \\ \hline

\end{tabular}
 \caption{Examples on which BERT is right but KIM is wrong.}
 \label{sample}
\end{table*}

\begin{table*}[t!]
\center
\renewcommand{\arraystretch}{1.5}
\small
\begin{tabular}{lll}
\hline
Example    &Sentences        &Label \\ \hline
6          &\begin{tabular}[c]{@{}l@{}} P: A woman wearing a patterned dress in an outdoor market sits surrounded by her offerings of \\ onions, eggs, \textbf{tomatoes}, beans, and many other things.\end{tabular}   & neutral \\
&         \begin{tabular}[c]{@{}l@{}} H: A woman wearing a patterned dress in an outdoor market sits surrounded by her offerings of \\ onions, eggs, \textbf{carrots}, beans, and many other things.\end{tabular}   &   \\ 
        \hline
                
7        & \begin{tabular}[c]{@{}l@{}} P: A woman wearing a patterned dress in an outdoor market sits surrounded by her offerings of \\ onions, eggs, \textbf{tomatoes}, beans, and many other things.\end{tabular}   & contradiction \\
 &        \begin{tabular}[c]{@{}l@{}} H: A woman wearing a patterned dress in an outdoor market sits surrounded by her offerings of \\ onions, eggs, \textbf{pumpkins}, beans, and many other things.\end{tabular}   &   \\ \hline

\end{tabular}
 \caption{Some sentence pairs with inconsistent true labels.}
 \label{Inconsistant}
\end{table*}

We can see in Table~\ref{matched} that the model using human edited knowledge (KIM) and those learned with unannotated data (e.g., BERT and GPT) complement each other. Along the diagonal, we can see BERT outperforms KIM and achieves the best performance.

It is also interesting to see that BERT and KIM are better than GPT on this task. This could be due to the use of sentence prediction objective in BERT, which may help capture word pair information between two sentences, while KIM explicitly incorporates different types of word-pair relations in WordNet.   

We have also performed the above experiment (Table~\ref{mismatched}) on the out-of-domain dataset. We observed similar results, except that on the out-of-domain test data, the performance of models are lower than that on the in-domain test data shown in Table~\ref{matched}.


\subsection{Case Study and Analysis}
We further performed detailed analyses on premise-hypothesis pairs where KIM is correct but the pretrained model (BERT) is wrong, and vice versa, on both the Glockner and  stress test set.

\paragraph{Only KIM Correct}
We found that for most cases in which the word-pair information required to make the judgment (e.g., antonym and synonym) does present in WordNet, KIM is more accurate than BERT, with an example shown in Table~\ref{KIM_sample}. 


\paragraph{Only BERT Correct}
We found when the pretrained model is correct but KIM wrong, the examples could be categorized into one of the following two situations.

\noindent \textit{Category {\uppercase\expandafter{\romannumeral1} }:  Word-pair information is missing in WordNet but learned by pretraining in BERT.}
As shown the example in {Table~\ref{sample}}, the word pair $\langle$ \textit{sun-lit}, \textit{moon-lit}$\rangle$ cannot be found in WordNet, but their relationship can be learned by BERT via leveraging the large corpora.




\noindent {\textit{Category {\uppercase\expandafter{\romannumeral2}}: Some word-pair relations can be found in WordNet but KIM does not make a correct decision.}}
In example 4 of {Table~\ref{sample}}, the word pair $\langle $\textit{north}, \textit{south}$\rangle$ as well as their relationship (co-hypernym) can be found in WordNet, but KIM did not categorize this example correctly. We found the attention did not give enough focus on these word pairs.



Note that in a number of cases in the Glockner dataset, the true labels are not consistent. As shown in Table~\ref{Inconsistant}, one of the two sentence pairs is labelled as ~\textit{neutral} and the other as 
~\textit{contradiction}. We excluded such cases in our analysis.




\section{Conclusions}
We run several experiments showing that the recent unsupervised-pretraining schema does help learn NLI related semantics to achieve better NLI prediction, including synonyms and hypernyms that are useful in judging entailment. The experiments also reveal that
external knowledge obtained from pretraining and human authorized resources complement each other, suggesting the potential benefit of combining these two approaches. 

\bibliography{ms}
\bibliographystyle{ms}

\end{document}